\documentclass{article}

\usepackage[nonatbib, preprint]{neurips_2020_ml4ad}

\usepackage[utf8]{inputenc} 
\usepackage[T1]{fontenc}    
\usepackage{hyperref}       
\usepackage{url}            
\usepackage{booktabs}       
\usepackage{amsfonts}       
\usepackage{nicefrac}       
\usepackage{microtype}      

\usepackage{amsmath}
\usepackage{multirow}
\usepackage{graphicx}
\usepackage{algorithm}
\usepackage{algpseudocode} 

\usepackage{float}
\newfloat{algorithm}{tbp}{lop}

\title{Affordance-based Reinforcement Learning \\ for
Urban Driving}

\author{
Tanmay Agarwal$^{*}$, Hitesh Arora$^{*}$, Jeff Schneider\\
$^*$Equal contribution\\
Robotics Institute, Carnegie Mellon University \\
\texttt{\{tanmaya, hiteshar, jeff.schneider\}@cs.cmu.edu}\\
}

\begin{document}

\maketitle

\begin{abstract}
Traditional autonomous vehicle pipelines that follow a modular approach have been very successful in the past both in academia and industry, which has led to autonomy deployed on road. Though this approach provides ease of interpretation, its generalizability to unseen environments is limited and hand-engineering of numerous parameters is required, especially in the prediction and planning systems. Recently, deep reinforcement learning has been shown to learn complex strategic games and perform challenging robotic tasks, which provides an appealing framework for learning to drive. 
In this work, we propose a deep reinforcement learning framework to learn optimal control policy using waypoints and low-dimensional visual representations, also known as affordances. We demonstrate that our agents when trained from scratch learn the tasks of lane-following, driving around intersections as well as stopping in front of other actors or traffic lights even in the dense traffic setting. We note that our method achieves comparable or better performance than the baseline methods on the original and NoCrash benchmarks on the CARLA simulator.
\end{abstract}
\vspace{-3pt}
\section{Introduction} \label{sec:introduction}

A recent survey conducted by the National Highway Traffic Safety Administration suggested that more than 94\% of road accidents in the U.S were caused by human errors \cite{singh2018critical}. This has led to the rapid evolution of the autonomous driving systems over the last several decades with the promise to prevent such accidents and improve the driving experience. But despite numerous research efforts in academia and industry, the autonomous driving problem remains a long-standing problem. This is because the current systems still face numerous real-world challenges including ensuring the accuracy and reliability of the prediction systems, maintaining the reasonability and optimality of the decision-making systems, to determining the safety and scalability of the entire system.

Another major factor that impedes success is the complexity of the problem that ranges from learning to navigate in constrained industrial settings, to learning to drive on highways, to navigation in dense urban environments. Navigation in dense urban environments requires understanding complex multi-agent dynamics including tracking multiple actors across scenes, predicting intent, and adjusting agent behavior conditioned on historical states. Additionally, the agent must be able to generalize to novel scenes and learn to take `sensible' actions for observations in the long tail of rare events \cite{GoogleWomanExample}. These factors provide a strong impetus for the need of general learning paradigms that are sufficiently `complex' to take these factors into consideration.

Traditionally, the autonomous driving approaches divide the system into modules and employ an array of sensors and algorithms to each of the different modules \cite{campbell2010autonomous, kanade1986autonomous, levinson2011towards, montemerlo2008junior, munir2018autonomous, wallace1985first, yurtsever2020survey}. 
Developing each of these modules makes the overall task much easier as each of the sub-tasks can independently be solved by popular approaches in the literature of computer vision \cite{forsyth2002computer, schalkoff1989digital}, robotics \cite{brady1982robot, laumond1998robot} and vehicle dynamics \cite{featherstone2000robot, rajamani2011vehicle}. But with the advent of deep learning \cite{lecun2015deep}, most of the current state-of-the-art systems use a variant of supervised learning over large datasets \cite{geiger2013vision} to learn individual components' tasks. The major disadvantage of these heavily engineered modular systems is that it is extremely hard to tune these subsystems and replicate the intended behavior which leads to its poor performance in new environments. 

Another approach is exploiting imitation learning where the aim is to learn a control policy for driving behaviors based on observations collected from expert demonstrations \cite{pomerleau1989alvinn, bojarski2017explaining, bojarski2016end, codevilla2019exploring, sauer2018conditional, codevilla2018end, muller2006off, rhinehart2018deep, zhang2016query, pan2017agile, silver2010learning, bansal2018chauffeurnet, chen2020learning}. The advantage of these methods is that the agent can be trained in an end-to-end fashion to learn the desired control behavior which significantly reduces the engineering effort of tuning each component that is common to more modular systems. However, the major drawback that these systems face are their scalability to novel situations.

More recently, deep reinforcement learning (DRL) has shown exemplary results towards solving sequential decision-making problems, including learning to play complex strategic games \cite{silver2016mastering, silver2017mastering, mnih2013playing, mnih2015human} , as well as completing complex robotic manipulation tasks \cite{gu2017deep, akkaya2019solving, zhang2015towards}. The superhuman performance attained using this learning paradigm motivates the question of whether it could be leveraged to solve the long-standing goal of creating autonomous vehicles. This approach of using reinforcement learning has inspired a few recent works \cite{Dosovitskiy2017CARLAAO, kendall2019learning, Liang2018CIRLCI, khan2019latent, kiran2020deep} that learn control policies for navigation task using high dimensional observations like images. The previous approach using DRL \cite{Dosovitskiy2017CARLAAO} reports poor performance on navigation tasks, while the imitation learning-based approach \cite{Liang2018CIRLCI} that achieves better performance, suffers from poor generalizability. 

Although learning policies from high dimensional state spaces remain challenging due to the poor sample complexity of most reinforcement learning (RL) algorithms, the strong theoretical formulation of reinforcement learning along with the nice structure of the autonomous driving problem, such as dense rewards and shorter time horizons, make it an appealing learning framework to be applied to autonomous driving. Moreover, RL also offers a data-driven corrective mechanism to improve the learned policies with the collection of increased data. These factors make us strongly believe in its potential to learn common urban driving skills for autonomous driving and form the basis for this work. Next, we present the main contributions of this work which can be summarized as follows.

\begin{enumerate}
    \item We propose using low-dimensional visual affordances and waypoints as navigation input to learn optimal control in the urban driving task. 
    \item We demonstrate that using a model-free on-policy reinforcement learning algorithm (Proximal policy optimization \cite{ppo}), our agents are able to learn performant driving policies in the CARLA \cite{Dosovitskiy2017CARLAAO} simulator that exhibit wide range of behaviours like lane-following, driving around intersections as well as stopping in front of other actors or traffic lights.
\end{enumerate}

The work is structured into the following sections. Sec.~\ref{sec:related_work} summarizes related methods in literature. Sec.~\ref{sec:method} describes our reinforcement learning formulation in detail along with the algorithm and training procedure described in Sec.~\ref{sec:dynamic_training}. We then present our set of experiments in Sec.~\ref{sec:dynamic_experiments} and discuss our results and conclusion in Sec.~\ref{sec:dynamic_results} and Sec.~\ref{sec:dynamic_discussion} respectively.

\vspace{-5pt}
\section{Related Work} \label{sec:related_work}

In this section, we review some of the popular methods in autonomous driving literature, broadly clustered into three common approaches: modular, imitation learning, and reinforcement learning.

\vspace{-4pt}
\subsection{Modular Approaches}

The modular approaches aim to divide the entire task into different sub-tasks and sub-modules that include the core functional blocks of localization and mapping, perception, prediction, planning, and decision making, and vehicle control \cite{campbell2010autonomous, kanade1986autonomous, levinson2011towards, montemerlo2008junior, munir2018autonomous, wallace1985first, yurtsever2020survey}. The localization and mapping subsystem senses the state of the world and locates the ego-vehicle with respect to the environment \cite{bresson2017simultaneous, kuutti2018survey, badue2019self, okuda2014survey, yurtsever2020survey}. This is followed by the perception sub-system that detects and tracks all surrounding static and dynamics objects \cite{arnold2019survey, okuda2014survey, petrovskaya2009model, yurtsever2020survey, desouza2002vision, badue2019self}. The intermediate representation produced by the perception subsystem then feeds into the prediction and planning subsystems that output an optimal plan of action \cite{schwarting2018planning, paden2016survey, badue2019self, okuda2014survey, yurtsever2020survey} based on the future trajectories of all the agents in the environment \cite{lefevre2014survey, djuric2018motion, mozaffari2019deep} and the planning cost function. Finally, the plan of action is mapped into low-level control actions that are responsible for the motor actuation. For more details on these approaches, we direct readers to \cite{yurtsever2020survey, badue2019self, levinson2011towards}. Although these systems offer high interpretability, they employ a heavily engineered approach that requires cumbersome parameter tuning and large amounts of annotated data to capture the diverse set of scenarios that the autonomous driving vehicle may face.

\vspace{-5pt}
\subsection{Imitation Learning}

Most imitation learning-based (IL) approaches aim to learn a control policy based on expert demonstrations in an end-to-end manner. Dating back to one of the earliest successful works on IL is ALVINN \cite{pomerleau1989alvinn} which uses a simple feedforward network to learn the task of the road following. Other popular works \cite{abbeel2010autonomous, syed2008game, ross2011reduction, ratliff2006maximum} have leveraged expert demonstrations either using supervised learning or formulating online imitation learning using game-theory or maximum margin classifiers. With the development of the deep learning \cite{lecun2015deep, krizhevsky2012imagenet}, the trend has shifted towards using CNNs where \cite{bojarski2016end} first demonstrated an end-to-end CNN architecture to predict steering angle directly from raw pixels. 
A major downside in the assumption of the above models is that the optimal action can be inferred solely from a single perception input. To that end, \cite{chi2017deep, yang2018end, xu2017end} propose to combine spatial and temporal cues using recurrent units to learn actions conditioned on the historical states and instantaneous camera observations. 

Another limitation is that IL policies still suffer in the closed loop testing as they cannot be controlled by human experts. This is because a vehicle trained to imitate the expert cannot be guided when it hits an intersection. \cite{codevilla2018end} proposes to condition the IL policies based on the high-level navigational command input which disambiguates the perceptuomotor mapping and allows the model to still respond to high-level navigational commands provided by humans or a mapping application. \cite{chen2020learning} decouples the sensorimotor learning task into two: learning to see and learning to act that involve learning two separate agents.
Although these methods have low sample complexity and can be robust with enough training data, their generalization ability to complicated environments is still questionable. This is because none of the above approaches reliably handle the dense traffic scenes and are prone to suffer from inherent dataset biases and lack of causality \cite{codevilla2019exploring}. Moreover, collecting a large amount of expert data remains an expensive process and is difficult to scale. 
These limitations restrict the usage of IL methods for large-scale learning of common urban driving behaviors.

\vspace{-5pt}
\subsection{Reinforcement Learning}

Since DRL is challenging to be applied in the real world primarily due to safety considerations and poor sample complexity of the state-of-art algorithms, most current research in the RL domain is increasingly being carried out on simulators, such as TORCS \cite{wymann2000torcs} and CARLA \cite{Dosovitskiy2017CARLAAO}, which can eventually be transferred to real world settings. 
The first work that demonstrated learning stable driving policies on TORCS was \cite{mnih2016asynchronous} that used A3C to learn a discrete action policy using only an RGB image. \cite{ddpg} extends the prior work to continuous action space by proposing DDPG, which also learns policies in an end-to-end fashion directly from raw pixels. \cite{kendall2019learning} demonstrated the use of the same algorithm to learn continuous-valued policy using a single monocular image as input in both simulated and real-world environments. 

The original CARLA work \cite{Dosovitskiy2017CARLAAO} released a new driving benchmark along with three baselines that used a modular, imitation learning, and reinforcement learning-based approach respectively. The RL baseline used the A3C algorithm \cite{mnih2016asynchronous} but reported poor results than the imitation learning one. These results were improved by \cite{Liang2018CIRLCI} that finetune the imitation learned agent using DDPG for continuous action space. Although the results are better than the original CARLA RL baseline \cite{Dosovitskiy2017CARLAAO}, this method relies heavily on the pre-trained imitation learned agent which makes it unclear whether the improvement comes from the RL fine-tuning. \cite{tanmaya2019learning} proposed to use low-dimensional navigational signal in the form of waypoints along with semantically segmented images to learn stable driving policy using Proximal Policy Optimization (PPO) \cite{ppo}. A major limitation of all the above models is that they do not handle behaviours around dense traffic or traffic-light at intersections. A concurrent work \cite{toromanoff2020end} aims to handle these behaviours and proposes an end-to-end trainable RL algorithm that predicts implicit affordances and uses them as the RL state to train a combination of Rainbow, IQN, and Ape-X \cite{hessel2018rainbow, dabney2018implicit, horgan2018distributed} algorithms. Although the work demonstrates impressive results, it learns a discrete-action policy and uses a heavily engineered approach for reward shaping which is different from our approach as defined in the next section. 

\section{Model-Free RL for Urban Driving} \label{sec:method}




The primary RL control task that we aim to learn here is the goal-directed urban driving task that includes subtasks such as  lane-following, driving around intersections, avoiding collision with other dynamic actors (vehicles and pedestrians) and following the traffic light rules. Building on the ideas presented in \cite{tanmaya2019learning, sauer2018conditional, toromanoff2020end}, we propose using waypoints as navigation input and affordances as state input, that encode the relevant state of the world in a simplified low-dimensional representation. These affordances can be predicted using a separate visual encoder that learns to predict attributes such as distance to the vehicle ahead, distance to the nearest stop sign, or nearest traffic light state from high-dimensional visual data. By dividing the urban driving task into the affordance prediction, and planning and control blocks, the difficulty of the latter block is reduced which can learn the optimal control policy based on low-dimensional affordances.


We evaluate three variants of our approach that are trained and evaluated on the CARLA\footnote{CARLA v0.9.6 - \href{https://carla.org/2019/07/12/release-0.9.6/}{https://carla.org/2019/07/12/release-0.9.6/}} urban driving simulator. The first variant, referred as $A$, assumes the affordances are provided from an external system (which is the simulator in our setup), and focuses on learning planning and control using RL. The second variant, referred as $I$,  trains a convolutional neural network (CNN) \cite{krizhevsky2012imagenet} to learn affordances implicitly in the intermediate representations that are crucial to determining the optimal control. We believe this to be our ultimate goal as it eliminates hand-engineering the affordances, enabling it to be scaled to diverse scenarios. Here, we follow the approach of \cite{tanmaya2019learning} to use a convolutional autoencoder to learn a latent representation from a stack of birds-eye-view (BEV) semantically segmented image inputs which is trained simultaneously and independently along with RL policy training. We also evaluate a third variant that is an intermediate step between the above two variants to understand the training performance and refer to it as $A+I$. This variant incorporates explicit affordances directly in the form low-dimensional representations as well as implicit affordances learnt using the convolutional autoencoder networks. These variants form the state space $\mathcal{S}$ of the Markov decision process (MDP) defined in the next section (Sec.~\ref{sec:dynamic_state_space}). Further, the action space $\mathcal{A}$ and reward function $\mathcal{R}$ that characterizes our formulation of the MDP are described in Sec.~\ref{sec:dynamic_action_space} and Sec.~\ref{sec:dynamic_reward} respectively. Our proposed RL formulation is then trained using Proximal policy optimization (PPO) \cite{ppo}, a state-of-the-art on-policy model-free RL algorithm.

\vspace{-4pt}
\subsection{State Space} \label{sec:dynamic_state_space}

We define the state space $\mathcal{S}$ to include sufficient environment information to enable the agent to optimally solve the navigation task with dynamic actors. For the navigation input, we use the waypoint features (${\mathbf{\tilde{w}}}$) as proposed in \cite{tanmaya2019learning} to help direct the agent to the target destination. As the waypoints encode the static routing features and do not encode the dynamic state of the world, we propose using low-dimensional affordances to encode dynamic actor and traffic light states. The \textit{dynamic obstacle affordance} ($\mathbf{\tilde{o}}$) encodes the distance and speed of the front dynamic actor. Using simulator information, we detect the obstacle in front of the driving agent within the obstacle-proximity-threshold distance (15m) and use the normalized distance and speed as the input such that its range is $[0.0, 1.0]$. If there is no obstacle within the obstacle-proximity-threshold, we set the input value as 1.0 to ensure monotonic input space. Similar, we use \textit{traffic light affordance} ($\mathbf{\tilde{t}}$) which includes the state and distance of the nearest traffic light that affects our agent obtained from the simulator.
    
We use additional inputs to capture the agent's position and motion. For position, we use normalized signed distance from the waypoint trajectory ($\tilde{n}$) to encode the position of the agent relative to the optimal trajectory. Also, we add distance to goal destination ($\tilde{g}$) as another input to enable the agent to learn optimal value function of states close to destination. To capture motion, we augment agent's current speed ($\tilde{v}$) and previous steer action ($\tilde{s}$) to input space.  Since the CARLA simulator does not provide the current steer value of the agent, we use the steer action ($\tilde{s}$) taken by the agent at the previous timestep as a proxy to encode steer state of the agent.

Thus, for the variant $A$, we use the state representation, $[\mathbf{\tilde{w}}.\mathbf{\tilde{o}}.\mathbf{\tilde{t}}.\tilde{n}.\tilde{v}.\tilde{s}.\tilde{g}]$ where the $(.)$ indicates the concatenation operator. We believe that this low-dimensional representation encodes all the sufficient information to solve the complex navigation tasks. For the variant $A+I$, we add the latent features $\mathbf{\tilde{h}}$ of an autoencoder that takes as input a stack of birds-eye-view (BEV) semantically segmented frames to generate an embedding encoding the affordances implicitly. It uses the state space, $[\mathbf{\tilde{h}}.\mathbf{\tilde{w}}.\mathbf{\tilde{o}}.\mathbf{\tilde{t}}.\tilde{n}.\tilde{v}.\tilde{s}.\tilde{g}]$, that concatenates the latent features with our earlier representation ($A$). The variant $I$ aims to explicitly remove the low-dimensional obstacle affordance $\mathbf{\tilde{o}}$ and uses the state  $[\mathbf{\tilde{h}}.\mathbf{\tilde{w}}.\mathbf{\tilde{t}}.\tilde{n}.\tilde{v}.\tilde{s}.\tilde{g}]$. Note that $A + I$ and $I$ still use the low-dimensional traffic light affordance explicitly.

\vspace{-4pt}
\subsection{Action Space} \label{sec:dynamic_action_space}

For our driving agent, it is natural to include continuous control actions of the steer $(s)$, throttle $(t)$, and brake $(b)$ in the action space $\mathcal{A}$ as they form the control input in the CARLA simulator. As the end control predictions may be noisier, we also reparameterize the throttle and brake actions in terms of a target speed set-point. Thus the predicted throttle and brake actions, denoted by $\hat{t}$ and $\hat{b}$ respectively, form the outputs of a classical PID controller \cite{rivera1986internal, aastrom1995pid} that attempts to match the set-point. This smoothens the control response as well as simplifies learning the continuous actions to just two actions, steer, and target speed. The predicted steer action, denoted by $\hat{s}$ lies in the range of $[-0.5, 0.5]$ that approximately maps to $[-40^{\circ}, 40^{\circ}]$ of steering angle whereas the predicted target speed, denoted by $\hat{v}$, lies in $[-1, 1]$ range and linearly maps to $[0, 20]$ km/h.

\vspace{-4pt}
\subsection{Reward Function} \label{sec:dynamic_reward}

We use a dense reward function $\mathcal{R}$ similar to the one proposed in \cite{tanmaya2019learning}, which constitutes of three different components. The \textit{Speed-based Reward ($R_s$)}, directly proportional to the current speed $u$ of the agent, incentivizes the agent to reach the goal destination in the shortest possible time. The \textit{Distance-based Penalty from Optimal Trajectory ($R_d$)}, directly proportional to the lateral distance $d$ between the centre of our agent and the optimal trajectory, incentivizes the agent to stay close to the planned optimal trajectory. The \textit{Infraction Penalty ($R_i$)}, activated by a collision or traffic light violation ($\mathbf{I}(i)$), penalizes our agent upon causing any infractions in the urban scenarios. The overall function along with each of the components can be mathematically defined by Eq.~\ref{eq:reward2}.
\vspace{-4pt}
\begin{equation}
\begin{split} \label{eq:reward2}
    \mathcal{R} = R_s + R_d & + \mathbf{I}(i) *  R_i \\
    R_s = \alpha * \mathrm{u}; R_d = - \beta * \mathrm{d} &; R_i = - \gamma * \mathrm{u} - \delta
\end{split}
\end{equation}



\vspace{-9pt}
\subsection{Training} \label{sec:dynamic_training}

\begin{figure}[t]
    \centering
    \includegraphics[width=0.7\textwidth]{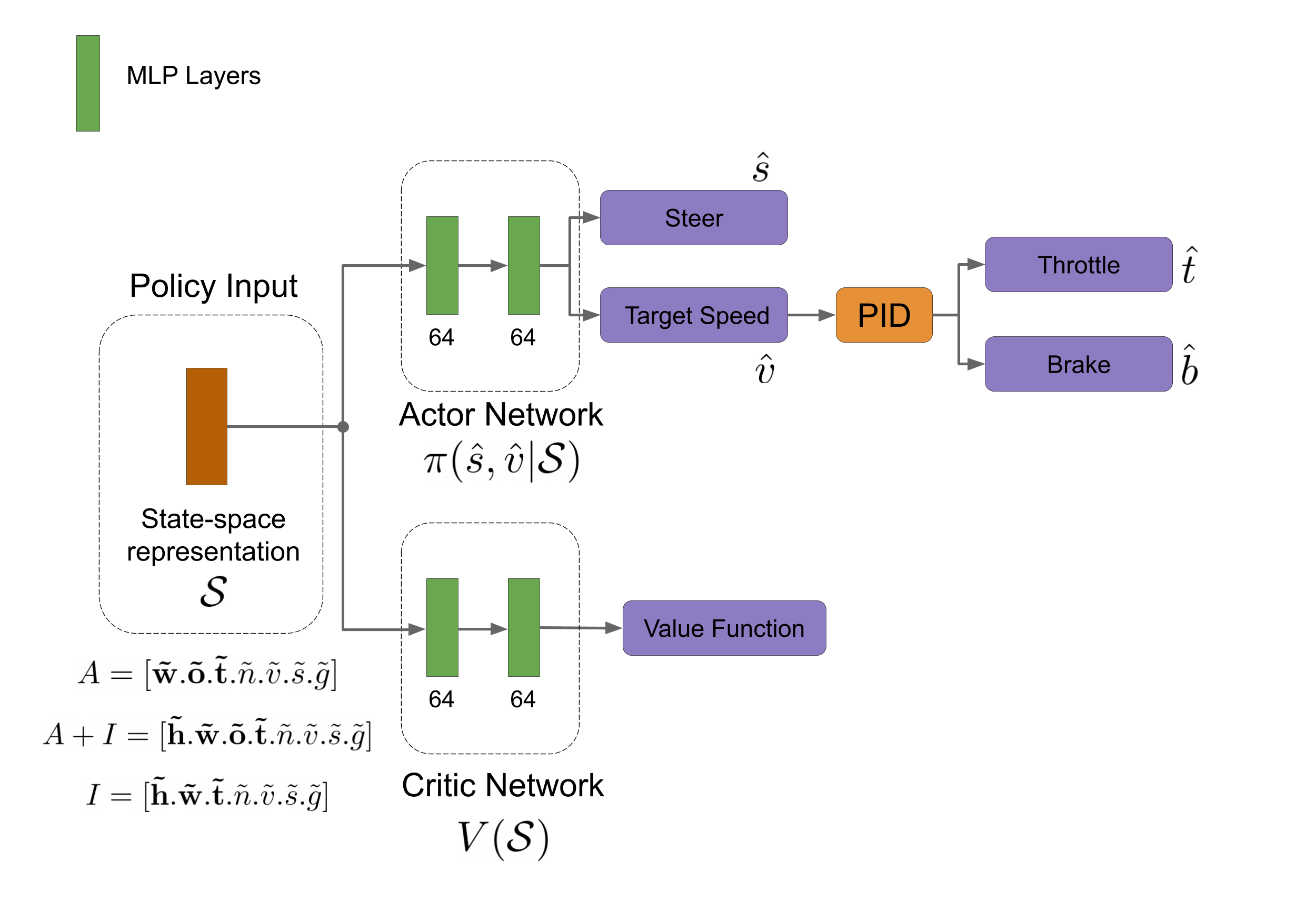}
    \caption{Overall architecture for the proposed approach. The state representation $S$ is a combination of waypoint features $\mathbf{\tilde{w}}$, dynamic obstacle affordance $\mathbf{\tilde{o}}$, traffic light affordance $\mathbf{\tilde{t}}$, previous time step steer and target speed actions ($\hat{s}, \hat{v}$), distance to goal destination $\hat{g}$, signed distance from the optimal trajectory $\hat{n}$ and latent features of the autoencoder $\mathbf{\tilde{h}}$ depending on the variant ($A$, $A+I$ or $I$). The policy network outputs the control actions $(\hat{s}, \hat{v})$ where $\hat{s}$ is the predicted steer and $\hat{v}$ is the predicted target speed which is then mapped to predicted throttle and brake $(\hat{t}, \hat{b})$ using a PID controller.}
    \label{fig:architecture2}
\end{figure}

We train the PPO algorithm by iteratively sampling a random episode with a different start and end destination. We then run an A* \cite{4082128} planner that computes a list of waypoints that trace the optimal trajectory from source to destination. The state representation we described earlier (Sec.~\ref{sec:dynamic_state_space}) is then queried at each time step from the simulator, which in real-world can be accessible from the perception and GPS subsystems. The agent steps through the simulator at every time step collecting on-policy rollouts that determine the updates to the policy and critic networks. The episode is terminated upon a collision, lane invasion, or traffic light infraction, or is deemed as a success if the agent reaches within distance $d$ m of the destination. This procedure repeats until the policy function approximator converges to the optimal policy. The actor and critic networks consist of a standard 2-layer feedforward network. All the networks are trained with a ReLU non-linearity and optimized using stochastic gradient descent with the Adam optimizer. The overall architecture of our approach is depicted in Fig.~\ref{fig:architecture2} and the overall algorithm is summarized in Algo.~\ref{algo:ppo_vae}.

\vspace{-6pt}
\section{Experiments} \label{sec:dynamic_experiments}

To implement our proposed algorithm (Algo.~\ref{algo:ppo_vae}), we build on top of stable-baselines implementation \cite{stable-baselines} and train our entire setup in the CARLA simulator \cite{Dosovitskiy2017CARLAAO}. The training is performed on the \textit{Dynamic Navigation} task of the original CARLA benchmark \cite{Dosovitskiy2017CARLAAO} and it is evaluated across all tasks of the same benchmark. Since this benchmark considers an episode to be successful regardless of any collisions or other infractions, we extend our evaluation to the recent NoCrash benchmark \cite{codevilla2019exploring} that proposes dynamic actor tasks that count collision as infractions. We then conduct a thorough infraction analysis that helps us analyze the different types of episode terminations and their relative proportions in order to understand the behavior and robustness of the policy learned by our agent.

\vspace{-5pt}
\subsection{Baselines}

For comparing our method with the prior work, we choose modular, imitation learning and reinforcement learning baselines that solve the goal-directed navigation task and report results on the original CARLA \cite{Dosovitskiy2017CARLAAO} and NoCrash \cite{codevilla2019exploring} benchmarks. The modular baselines include \textit{MP} \cite{Dosovitskiy2017CARLAAO}, \textit{CAL} \cite{sauer2018conditional} and \textit{AT}, which is the automatic control agent available within CARLA distribution. \textit{IL} \cite{Dosovitskiy2017CARLAAO}, \textit{CIL} \cite{codevilla2018end}, \textit{CILRS} \cite{codevilla2019exploring} and \textit{LBC} \cite{chen2020learning} form our imitation learning baselines whereas on the reinforcement learning end we choose \textit{RL} \cite{Dosovitskiy2017CARLAAO}, \textit{CIRL} \cite{Liang2018CIRLCI} and \textit{IA} \cite{toromanoff2020end}. We point readers to refer the Appendix for a brief summary of each of our chosen baselines. Although all of the above baselines use forward-facing RGB image and high-level navigational command as inputs, we recognize the differences in our inputs and presume that the current state-of-the-art perception systems are capable of predicting the low-dimensional representations or semantic labels with reasonable accuracy. Additionally, we do not consider the pedestrian actors in our setup owing to the limitations of CARLA 0.9.6 version.





\vspace{-9pt}
\section{Results} \label{sec:dynamic_results}

\begin{figure}[t]
    \centering
    \includegraphics[width=\textwidth]{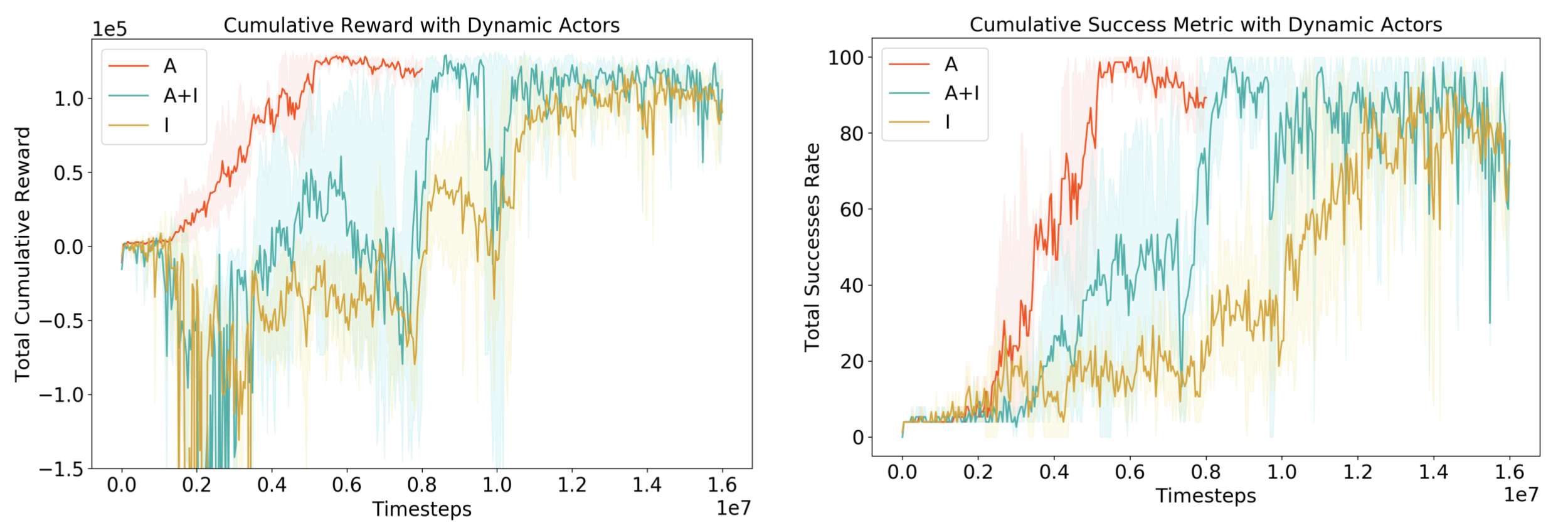}
    \caption{The figure reports the mean cumulative reward and success rate for the three variants: $A, A+I \text{ and } I$ on the \textit{Dynamic Navigation} task \cite{Dosovitskiy2017CARLAAO}. The plots indicate that the $A$ and $A+I$ learn the navigation task successfully whereas the state-representation $I$ learns the task slowly as observed by the performance improvement after 12M time steps of training. Shaded region corresponds to the minimum and maximum values across 3 different seeds.}
    \label{fig:dynamic_results}
\end{figure}

Comparing the training stability across our three variants ($A$, $A+I$, or $I$) as defined in Sec.~\ref{sec:method}, we observe from Fig.~\ref{fig:dynamic_results} that the variant $A$ that explicitly uses affordances, performs the best as it is stably able to learn the dynamic navigation task on the original CARLA benchmark \cite{Dosovitskiy2017CARLAAO} within 6M timesteps. We also see that the variants $I$ and $A + I$, take almost twice the timesteps (12M) to reach similar performance. $I$ is slower compared with the other two variants, which makes us empirically believe that it requires more samples to learn the dynamic obstacle affordance when compared with the latter that explicitly encode it.

Considering our input representation includes low-dimensional representations or latent features of semantically segmented images, we compare and present results only on training weather conditions from all the baselines. For evaluation, we pick the best performing model from each seed based on the cumulative reward it collects at the time of validation and report results averaged over 3 seeds and 5 different evaluations of the benchmarks.

\vspace{-7pt}
\subsection{Evaluation on the Original CARLA Benchmark}

On the original CARLA benchmark \cite{Dosovitskiy2017CARLAAO}, we note that both the variants $A$ and $A+I$ (Table.~\ref{table:low_dim_original_benchmark_baseline}) achieve a perfect success percentage on all the driving tasks in both \textit{Town 01} and \textit{Town 02}, whereas the variant $I$ achieves near-perfect success percentage. The driving policy learned by our agent demonstrates stopping in front of other actors or traffic lights in red-state while perfectly navigating through the town and around intersections. Further, for all the variants, our results also demonstrate that on the most difficult driving task of \textit{Dynamic Navigation}, we achieve a significant improvement in the success rate performance on both towns when compared with our baselines such as \textit{MP} \cite{Dosovitskiy2017CARLAAO}, \textit{IL} \cite{Dosovitskiy2017CARLAAO}, \textit{RL} \cite{Dosovitskiy2017CARLAAO}, \textit{CIL} \cite{codevilla2018end}, \textit{CIRL} \cite{Liang2018CIRLCI} and \textit{CAL} \cite{sauer2018conditional}. For \textit{CILRS} \cite{codevilla2019exploring}, the improvement is significant for \textit{Town 02} and moderate for \textit{Town 01}. We also note that our method reports comparable performance to the recent works of \textit{LBC} \cite{chen2020learning} and \textit{IA} \cite{toromanoff2020end} on some of the tasks.

\vspace{-4pt}
\subsection{Evaluation on the NoCrash Benchmark}

\vspace*{-5pt}
\begin{table}[]
\centering
\resizebox{\textwidth}{!}{
\begin{tabular}{cccccccccc}
\toprule
\multicolumn{10}{c}{\textbf{NoCrash Benchmark (\% Success Episodes)}} \\ \hline
\multicolumn{1}{c}{\textbf{Task}} & \multicolumn{9}{c}{\textbf{Training Conditions (\textit{Town 01})}} \\ \hline
\multicolumn{1}{c}{} & \textit{CIL} & \textit{CAL} & \textit{CILRS} & \textit{LBC} & \textit{IA} & \textit{AT} & \textit{Ours (A)} & \textit{Ours (A+I)} & \textit{Ours (I)} \\
\multicolumn{1}{c}{\textit{Empty}} & $79 \pm 1$ & $81 \pm 1$ & $87 \pm 1$ & $97 \pm 1$ & \textbf{100} & $\textbf{100} \pm 0$ & $\textbf{100} \pm 0$ & $\textbf{100} \pm 0$ & $94 \pm 10$ \\
\multicolumn{1}{c}{\textit{Regular}} & $60 \pm 1$ & $73 \pm 2$ & $83 \pm 0$ & $93 \pm 1$ & 96 & $\textbf{99} \pm 1$ & $98 \pm 2$ & $90 \pm 2$ & $90 \pm 4$ \\
\multicolumn{1}{c}{\textit{Dense}} & $21 \pm 2$ & $42 \pm 1$ & $42 \pm 2$ & $71 \pm 5$ & 70 & $86 \pm 3$ & $\textbf{95} \pm 2$ & $94 \pm 4$ & $89 \pm 4$ \\
\hline
\multicolumn{1}{c}{\textbf{Task}} & \multicolumn{9}{c}{\textbf{Testing Conditions (\textit{Town 02})}} \\ \hline
\multicolumn{1}{c}{} & \textit{CIL} & \textit{CAL} & \textit{CILRS} & \textit{LBC} & \textit{IA} & \textit{AT} & \textit{Ours (A)} & \textit{Ours (A+I)} & \textit{Ours (I)} \\
\multicolumn{1}{c}{\textit{Empty}} & $48 \pm 3$ & $36 \pm 6$ & $51 \pm 1$ & $\textbf{100} \pm 0$ & 99 & $\textbf{100} \pm 0$ & $\textbf{100} \pm 0$ & $\textbf{100} \pm 0$ & $98 \pm 3$ \\
\multicolumn{1}{c}{\textit{Regular}} & $27 \pm 1$ & $26 \pm 2$ & $44 \pm 5$ & $94 \pm 3$ & 87 & $\textbf{99} \pm 1$ & $98 \pm 1$ & $96 \pm 2$ & $96 \pm 2$ \\
\multicolumn{1}{c}{\textit{Dense}} & $10 \pm 2$ & $9 \pm 1$ & $38 \pm 2$ & $51 \pm 3$ & 42 & $60 \pm 3$ & $\textbf{91} \pm 1$ & $89 \pm 2$ & $79 \pm 3$ \\
\bottomrule
\end{tabular}}
\caption{Quantitative comparison with the baselines on the NoCrash benchmark \cite{codevilla2019exploring}. The table reports the percentage (\%) of successfully completed episodes for each task in the training and testing town. \textbf{Higher} is better. The baselines include \textit{CIL} \cite{codevilla2018end}, \textit{CAL} \cite{sauer2018conditional}, \textit{CILRS} \cite{codevilla2019exploring}, \textit{LBC} \cite{chen2020learning}, \textit{IA} \cite{toromanoff2020end} and CARLA built-in autopilot control (\textit{AT}) compared with our PPO method. Results denote average of 3 seeds across 5 trials. \textbf{Bold} values correspond to the best mean success rate.}
\label{table:low_dim_nocrash_benchmark_baseline}
\end{table}

On the NoCrash benchmark \cite{codevilla2019exploring}, our quantitative results (Table~\ref{table:low_dim_nocrash_benchmark_baseline}) again demonstrate that our driving agent learns the optimal control almost perfectly across different levels of traffic and town conditions. The variant  $A$ outperforms the variant $A+I$ which outperforms variant $I$. This difference can be attributed to the low-dimensionality of the state representation $A$. Moreover, we also note that the success rate performance achieved by our agent is significantly higher than all the prior modular and imitation learning baselines such as \textit{CIL} \cite{codevilla2018end}, \textit{CAL} \cite{sauer2018conditional}, \textit{CILRS} \cite{codevilla2019exploring}, \textit{LBC} \cite{chen2020learning} and \textit{IA} \cite{toromanoff2020end}. The \textit{AT} baseline, which refers to the autopilot control that is shipped with the CARLA binaries, uses a hand-engineered approach to determine the optimal control. We observe that even on the hardest task of \textit{Dense} traffic, our method using all three variants significantly outperforms even the most engineered approach to urban driving. For a qualitative analysis, we publish a video\footnote{\href{https://www.youtube.com/watch?v=AwbJSPtKHkY}{https://www.youtube.com/watch?v=AwbJSPtKHkY}} of our agent successfully driving on the dynamic navigation task. We presume that our near-perfect results can be accounted for our simpler choice of input representation. Thus for a fair comparison, our future direction is to move towards high-dimensional RGB images. Nevertheless, a notable difference between the baseline methods and our method is that they use strong supervision signals, unlike our method that learns the optimal policy from scratch based on trial-and-error learning. These supervised signals, often expensive to collect, do not always capture the distribution of scenarios entirely. In contrast, we show that our results demonstrate the prospects of utilizing reinforcement learning to learn complex urban driving behaviors.

\vspace{-7pt}
\subsection{Infraction Analysis}

\begin{table}[t]
\centering
\resizebox{\textwidth}{!}{
\begin{tabular}{cccccccccc}
\toprule
\multicolumn{10}{c}{\textbf{Infraction Analysis on NoCrash Benchmark (\% Episodes)}} \\ \hline
\textbf{Task} & \textbf{Metric} & \multicolumn{4}{c}{\textbf{Training Conditions (\textit{Town 01})}} & \multicolumn{4}{c}{\textbf{Testing Conditions (\textit{Town 02})}} \\ \hline
\multicolumn{2}{l}{} & \textit{CIL} & \textit{CAL} & \textit{CILRS} & \textit{Ours (A)} & \textit{CIL} & \textit{CAL} & \textit{CILRS} & \textit{Ours (A)} \\ \hline
\multirow{4}{*}{Empty} & Success & 79.00 & 84.00 & 96.33 & \textbf{100.00} & 41.67 & 48.67 & 72.33 & \textbf{100.00} \\
& Col. Vehicles & \textbf{0.00} & \textbf{0.00} & \textbf{0.00} & \textbf{0.00} & \textbf{0.00} & \textbf{0.00} & \textbf{0.00} & \textbf{0.00} \\
& Col. Other & 11.00 & 9.00 & 1.33 & \textbf{0.00} & 51.00 & 45.33 & 20.00 & \textbf{0.00} \\
& Timeout & 10.00 & 7.00 & 2.33 & \textbf{0.00} & 7.33 & 6.00 & 7.67 & \textbf{0.00} \\
\hline
\multirow{4}{*}{Regular} & Success & 61.50 & 57.00 & 87.33 & \textbf{98.40} & 22.00 & 27.67 & 49.00 & \textbf{98.16} \\
& Col. Vehicles & 16.00 & 26.00 & 4.00 & \textbf{0.27} & 34.67 & 30.00 & 12.67 & \textbf{0.00} \\
& Col. Other & 16.50 & 14.00 & 5.67 & \textbf{0.53} & 37.33 & 36.33 & 28.00 & \textbf{0.92} \\
& Timeout & 6.00 & 3.00 & 3.00 & \textbf{0.80} & 6.00 & 6.00 & 10.33 & \textbf{0.92} \\
\hline
\multirow{4}{*}{Dense} & Success & 22.00 & 16.00 & 41.66 & \textbf{95.38} & 7.33 & 10.67 & 21.00 & \textbf{91.20} \\
& Col. Vehicles & 49.50 & 57.00 & 20.67 & \textbf{0.62} & 55.67 & 46.33 & 35.00 & \textbf{3.73} \\
& Col. Other & 25.00 & 24.00 & 34.67 & \textbf{1.23} & 34.33 & 35.33 & 35.00 & \textbf{1.87} \\
& Timeout & 3.50 & 3.00 & 3.00 & \textbf{2.77} & \textbf{2.67} & 7.67 & 9.00 & 3.20 \\ \bottomrule   
\end{tabular}}
\caption{Quantitative analysis of episode termination causes and comparison with the baselines on the NoCrash benchmark \cite{codevilla2019exploring}. The table reports the percentage (\%) of episodes with their termination causes for each task in the training and testing town. The columns for a single method/task/condition should add up to 1. \textbf{Bold} values correspond to the \textbf{best} performance for each termination condition. Results denote average of 3 seeds across 5 trials.}
\label{table:low_dim_nocrash_infraction_analysis}
\end{table}

To analyze the failure cases of our agent, we also perform an infraction analysis on variant $A$ that reports the percentage of episodes for each termination condition and driving task on the NoCrash benchmark \cite{codevilla2019exploring}. We then compare this analysis with few of our baselines like \textit{CIL} \cite{codevilla2018end}, \textit{CAL} \cite{sauer2018conditional} and \textit{CILRS} \cite{codevilla2019exploring} that perform end-to-end imitation learning or take a modular approach to predicting low-dimensional affordances like ours. We incorporate the baseline infraction metrics from the \textit{CILRS} work \cite{codevilla2019exploring}. We observe from this analysis (Table~\ref{table:low_dim_nocrash_infraction_analysis}) that our agent significantly outperforms all the other baselines across all traffic and town conditions. Further, we note that our agent in the \textit{Empty} town task achieves a perfect success percentage across both \textit{Town 01} and \textit{Town 02} without facing any infractions. Additionally, on the \textit{Regular} and \textit{Dense} traffic tasks, we notice that our approach reduces the number of collision and timeout infractions by at least an order of magnitude across both the towns. Therefore, our results indicate that the policy learned by our agents using reinforcement learning is robust to variability in both traffic and town conditions.

\vspace{-8pt}
\section{Conclusion} \label{sec:dynamic_discussion}

In this work, we present an approach to learn urban driving tasks that commonly subtasks such as lane-following, driving around intersections and handling numerous interactions between dynamic actors, and traffic signs and signals. We formulate a reinforcement learning-based approach primarily based around three different variants that use waypoints and low-dimensional visual affordances. We demonstrate that using such low-dimensional representations makes the planning and control problem easier as we can learn stable and robust policies demonstrated by our results with state representation $A$. Further, we also observe that as we move towards learning these representations inherently using convolutional encoders, the performance and robustness of our learned policy decreases which requires more training samples to learn the optimal representations, as evident from Fig.~\ref{fig:dynamic_results} and the results presented in Sec.~\ref{sec:dynamic_results}. We also note that our method when trained from scratch achieves comparable or better performance than most baselines methods that often require expert supervision to learn control policies. We also show that our agent learned with variant $A$ reports a significantly lower number of infractions that are at least an order of magnitude less than the prior works on the same tasks. Thus, our work shows the strong potential of using reinforcement learning for learning autonomous driving behaviours and we hope it inspires more research in the future.

\small

\bibliographystyle{plain}
\bibliography{references}

\normalsize

\newpage
\section*{Appendix}

\section*{Algorithm}

\begin{algorithm}[h]
\begin{algorithmic}[1]
\State Input: initial policy parameters $\theta_{0},$ initial value function parameters $\phi_{0}$, pretrained auto-encoder parameters $\beta_{0},$
\For{$k=0,1,2, \ldots$}
\State Collect trajectories $\mathcal{D}_{k}=\left\{\tau_{i}\right\}$ by running policy $\pi_{\theta_{k}}$ in the 
environment.

\State Compute rewards-to-go $\hat{R}_{t}$.
\State Compute advantage estimates, $\hat{A}_{t}$ based on the current value function $V_{\phi_{k}}$.
\State Update the policy by maximizing the PPO-Clip objective.

\[
\theta_{k+1}=\arg \max _{\theta} \frac{1}{\left|\mathcal{D}_{k}\right| T} \sum_{\tau \in \mathcal{D}_{k}} \sum_{t=0}^{T} \min \bigg( r(\theta)  A_{\pi_{\theta_{k}}}(s_t, a_t), \text{clip}\Big(r(\theta), 1 - \epsilon, 1 + \epsilon \Big)A_{\pi_{\theta_{k}}}(s_t, a_t) \bigg)
\]

\State Fit value function by regression on mean-squared error.
\[
\phi_{k+1}=\arg \min _{\phi} \frac{1}{\left|\mathcal{D}_{k}\right| T} \sum_{\tau \in \mathcal{D}_{k}} \sum_{t=0}^{T}\left(V_{\phi}\left(s_{t}\right)-\hat{R}_{t}\right)^{2}
\]

\State Every fixed $n$ steps, finetune the autoencoder based on cross-entropy loss.

\[
\beta_{k+1}=\arg \min _{\beta} \sum_{p \in \mathbf{SS_\text{image}}} \sum_{c \in \mathcal{C}} \left(- t_c(p) \log o_c(p) \right)
\]

\EndFor
\end{algorithmic}
\caption[PPO+AE]{Learning to Drive via Model-Free RL on Learned Representations}
\label{algo:ppo_vae}
\end{algorithm}

\section*{Summary of Baselines} \label{sec:appendix_baselines_summary}
We compare our work with the following baselines that solve the goal-directed navigation task using an either modular approach, end-to-end imitation learning, or reinforcement learning. Since most of the works do not have open-source implementations available or report results on the older versions of CARLA, we report the numbers directly from their work. 
\begin{itemize}
    \item CARLA \textit{MP}, \textit{IL} \& \textit{RL} \cite{Dosovitskiy2017CARLAAO}: These baselines, proposed in the original CARLA work \cite{Dosovitskiy2017CARLAAO} suggest three different approaches to the autonomous driving task. The modular pipeline (MP) uses a vision-based module, a rule-based planner, and a classical controller. The imitation learning (IL) one learns a deep network that maps sensory input to driving commands whereas the reinforcement learning (RL) baseline does end-to-end RL using the A3C algorithm \cite{mnih2016asynchronous}. 
    \item \textit{AT}: This baseline refers to the CARLA built-in autopilot control that uses a hand-engineered approach to determine optimal control.
    \item \textit{CIL} \cite{codevilla2018end}: This work proposes a conditional imitation learning pipeline that learns a driving policy from expert demonstrations of low-level control inputs, conditioned on the high-level navigational command.
    \item \textit{CIRL} \cite{Liang2018CIRLCI}: This work proposes to use a pre-trained imitation learned policy to carry-out off-policy reinforcement learning using the DDPG algorithm \cite{ddpg}.
    \item \textit{CAL} \cite{sauer2018conditional}: This baseline proposes to learn a separate visual encoder that predicts low-dimensional representations, also known as affordances, that are essential for urban driving. These representations are then fused with classical controllers. 
    \item \textit{CILRS} \cite{codevilla2019exploring}: This work builds on top of CIL \cite{codevilla2018end} to propose a robust behavior cloning pipeline that generalizes well to complex driving scenarios. The method suggests learning a ResNet architecture \cite{he2016deep} that predicts the desired control as well as the agent's speed to learn speed-related features from visual cues. 
    \item \textit{LBC} \cite{chen2020learning}: This work decouples the sensorimotor learning task into two, learning to see and learning to act. In the first step, a privileged agent is learned that has access to the simulator states and learns to act. The second step involves learning to see that learns an agent based on supervision provided by the privileged agent.
    \item \textit{IA} \cite{toromanoff2020end}: This work proposes to learn a ResNet encoder \cite{he2016deep} that predicts the implicit affordances and uses its output features to learn a separate policy network optimized using DQN algorithm \cite{mnih2013playing}. 
\end{itemize}

\section*{Evaluation on the Original CARLA Benchmark}

On the original CARLA benchmark \cite{Dosovitskiy2017CARLAAO}, we note that both the variants $A$ and $A+I$ (Table.~\ref{table:low_dim_original_benchmark_baseline}) achieve a perfect success percentage on all the driving tasks in both \textit{Town 01} and \textit{Town 02}, whereas the variant $I$ achieves near-perfect success percentage. 

\begin{table}[h]
\centering
\resizebox{0.9\textwidth}{!}{
\begin{tabular}{ccccccccccccc}
\toprule
\multicolumn{13}{c}{\text{\textbf{Original CARLA Benchmark (\% Success Episodes)}}} \\ \hline
\multicolumn{1}{c}{\textbf{Task}} & \multicolumn{12}{c}{\textbf{Training Conditions (\textit{Town 01})}} \\ \hline
\multicolumn{1}{c}{} & \textit{MP} & \textit{IL} & \textit{RL} & \textit{CIL} & \textit{CIRL} & \textit{CAL} & \textit{CILRS} & \textit{LBC} & \textit{IA} & \textit{Ours (A)} & \textit{Ours (A+I)} & \textit{Ours (I)} \\
\multicolumn{1}{c}{\textit{Straight}} & 98 & 95 & 89 & 98 & 98 & \textbf{100} & 96 & \textbf{100} & \textbf{100} & $\textbf{100} \pm 0$ & $\textbf{100} \pm 0$ & $99 \pm 2$ \\
\multicolumn{1}{c}{\textit{One Turn}} & 82 & 89 & 34 & 89 & 97 & 97 & 92 & \textbf{100} & \textbf{100} & $\textbf{100} \pm 0$ & $\textbf{100} \pm 0$ & $\textbf{100} \pm 0$ \\
\multicolumn{1}{c}{\textit{Navigation}} & 80 & 86 & 14 & 86 & 93 & 92 & 95 & \textbf{100} & \textbf{100} & $\textbf{100} \pm 0$ & $\textbf{100} \pm 0$ & $96 \pm 7$ \\
\multicolumn{1}{c}{\textit{Dyn. Navigation}} & 77 & 83 & 7 & 83 & 82 & 83 & 92 & \textbf{100} & \textbf{100} & $\textbf{100} \pm 0$ & $\textbf{100} \pm 0$ & $95 \pm 7$ \\
\multicolumn{1}{c}{} & \multicolumn{12}{c}{} \\ \hline
\multicolumn{1}{c}{\textbf{Task}} & \multicolumn{12}{c}{\textbf{Testing Conditions (\textit{Town 02})}} \\ \hline
\multicolumn{1}{c}{} & \textit{MP} & \textit{IL} & \textit{RL} & \textit{CIL} & \textit{CIRL} & \textit{CAL} & \textit{CILRS} & \textit{LBC} & \textit{IA} & \textit{Ours (A)} & \textit{Ours (A+I)} & \textit{Ours (I)} \\
\multicolumn{1}{c}{\textit{Straight}} & 92 & 97 & 74 & 97 & \textbf{100} & 93 & 96 & \textbf{100} & \textbf{100} & $\textbf{100} \pm 0$ & $\textbf{100} \pm 0$ & $\textbf{100} \pm 0$ \\
\multicolumn{1}{c}{\textit{One Turn}} & 61 & 59 & 12 & 59 & 71 & 82 & 84 & \textbf{100} & \textbf{100} & $\textbf{100} \pm 0$ & $\textbf{100} \pm 0$ & $97 \pm 2$ \\
\multicolumn{1}{c}{\textit{Navigation}} & 24 & 40 & 3 & 40 & 53 & 70 & 69 & 98 & \textbf{100} & $\textbf{100} \pm 0$ & $99 \pm 1$ & $\textbf{100} \pm 0$ \\
\multicolumn{1}{c}{\textit{Dyn. Navigation}} & 24 & 38 & 2 & 38 & 41 & 64 & 66 & 99 & 98 & $\textbf{100} \pm 0$ & $\textbf{100} \pm 0$ & $99 \pm 1$ \\
\multicolumn{1}{c}{} & \multicolumn{12}{c}{} \\ \bottomrule
\end{tabular}}
\caption{Quantitative comparison with the baselines that solve the four goal-directed navigation tasks using modular, imitation learning or reinforcement learning approaches on the original CARLA benchmark \cite{Dosovitskiy2017CARLAAO}. The table reports the percentage (\%) of successfully completed episodes for each task in the training (\textit{Town 01}) and testing town (\textit{Town 02}). \textbf{Higher} is better. The baselines include \textit{MP} \cite{Dosovitskiy2017CARLAAO}, \textit{IL} \cite{Dosovitskiy2017CARLAAO}, \textit{RL} \cite{Dosovitskiy2017CARLAAO}, \textit{CIL} \cite{codevilla2018end}, \textit{CIRL} \cite{Liang2018CIRLCI}, \textit{CAL} \cite{sauer2018conditional}, \textit{CILRS} \cite{codevilla2019exploring}, \textit{LBC} \cite{chen2020learning} and \textit{IA} \cite{toromanoff2020end} compared with our PPO method. Results denote average of 3 seeds across 5 trials. \textbf{Bold} values correspond to the best mean success rate.}
\label{table:low_dim_original_benchmark_baseline}
\end{table}

\section*{Hyperparameters}

\subsection*{CARLA Environment}

In this subsection, we detail all the parameters we use for our experiments with the CARLA simulator \cite{Dosovitskiy2017CARLAAO}. The list mentioned below covers all the parameters to the best of our knowledge, except those that are set to default values in the simulator.

\begin{itemize}
    \item Camera top-down co-ordinates: ($x=13.0, y=0.0, z=18.0, pitch=270^{\circ}$)
    \item Camera front-facing co-ordinates: ($x=2.0, y=0.0, z=1.4, pitch=0^{\circ}$)
    \item Camera image resolution: ($x=128, y=128$)
    \item Camera field-of-view = 90
    \item Server frame-rate = $10fps$
    \item Maximum target-speed = $20km/h$
    \item PID parameters: ($K_P = 0.1, K_D = 0.0005, K_I = 0.4, dt = 1/10.0$)
    \item Waypoint resolution ($w_d$) = $2m$
    \item Number of next waypoints ($n$) = 5
    \item Maximum time steps ($m$) = 10000
    \item Success distance from goal ($d$) = $10m$
\end{itemize}


\subsection*{Algorithm Hyperparameters}

In this subsection, we detail all the algorithm hyper-parameters used for the experiments and results reported in Sec.~\ref{sec:dynamic_results}. The list mentioned below covers all the parameters to the best of our knowledge, except those that are set to default values as defined in the stable baselines documentation \cite{stable-baselines}.

\begin{itemize}
    \item Total training time steps = 16M
    \item N-steps = 10000
    \item Number of epochs = 10
    \item Number of minibatches = 20
    \item Clip parameter = 0.1
    \item Speed-based reward coefficient ($\alpha$) = 1
    \item Distance-based penalty from optimal trajectory ($\beta$) = 1
    \item Infraction penalty speed-based coefficient ($\gamma$) = 250
    \item Infraction penalty constant coefficient ($\delta$) = 250
    \item Learning rate = 0.0002
    \item Validation interval = 40K
    \item Number of dynamic actors at training time = $\mathcal{U}(70, 150)$, where $\mathcal{U}$ refers to uniform distribution.
    \item Image frame-stack fed to autoencoder ($k$) = 3
    \item Dynamic obstacle proximity threshold ($d_{prox}$) = $15m$
    \item Traffic light proximity threshold ($t_{prox}$) = $15m$
    \item Minimum threshold distance for traffic light detection = $6m$
    \item Random seeds = 3
    \item Number of benchmark evaluations = 5
\end{itemize}

\end{document}